\documentclass[journal]{IEEEtran}
\usepackage{amsmath,amsfonts}
\usepackage{algorithm2e}
\usepackage{array}
\usepackage{textcomp}
\usepackage{stfloats}
\usepackage{url}
\usepackage{verbatim}
\usepackage{graphicx}
\usepackage{adjustbox}
\usepackage{cite}
\usepackage{hyperref}
\usepackage{url}
\usepackage{booktabs}
\usepackage{braket} 
\usepackage{soul}
\usepackage{placeins}
\usepackage[dvipsnames]{xcolor}
\usepackage{colortbl} 

\SetKwComment{Comment}{/* }{ */}
\RestyleAlgo{ruled}

\usepackage{xcolor} 

\usepackage[switch]{lineno}
\usepackage{multicol}
\usepackage{multirow}
\usepackage{mwe}
\usepackage{float}
\usepackage{balance}
\usepackage{lipsum}
\usepackage{balance}

\ifCLASSINFOpdf
\else
\fi
\begin{document}
%
\title{A Deep Learning framework for building damage assessment using VHR SAR and geospatial data: demonstration on the 2023 Türkiye Earthquake}
%
%
%

\author{
    \IEEEauthorblockN{Luigi Russo, \IEEEmembership{Student Member, IEEE}, 
    Deodato Tapete, \IEEEmembership{Member, IEEE}, 
    Silvia Liberata Ullo, \IEEEmembership{Senior Member, IEEE}, 
    and Paolo Gamba, \IEEEmembership{Fellow, IEEE}}
    
    \thanks{Luigi Russo and Paolo Gamba are with the Department of Electrical, Computer and Biomedical Engineering, University of Pavia, 27100 Pavia, Italy (e-mail: luigi.russo02@universitadipavia.it; paolo.gamba@unipv.it).}

    \thanks{Silvia Liberata Ullo is with the Department of Engineering, University of Sannio, 82100 Benevento, Italy (e-mail: ullo@unisannio.it).}

    \thanks{Deodato Tapete is with the Italian Space Agency (ASI), Via del Politecnico, 00133 Roma RM, Italia (e-mail: deodato.tapete@asi.it).}
}

\maketitle
 
\begin{abstract}
Building damage identification shortly after a disaster is crucial for guiding emergency response and recovery efforts. Although optical satellite imagery is commonly used for disaster mapping, its effectiveness is often hampered by cloud cover or the absence of pre-event acquisitions. To overcome these challenges, we introduce a novel multimodal deep learning (DL) framework for detecting building damage using single-date very high resolution (VHR) Synthetic Aperture Radar (SAR) imagery from the Italian Space Agency’s (ASI) COSMO-SkyMed (CSK) constellation, complemented by auxiliary geospatial data. Our method integrates SAR image patches, OpenStreetMap (OSM) building footprints, digital surface model (DSM) data, and structural and exposure attributes from the Global Earthquake Model (GEM) to improve detection accuracy and contextual interpretation.
Unlike existing approaches that depend on pre- and post-event imagery, our model utilizes only post-event data, facilitating rapid deployment in critical scenarios. The framework’s effectiveness is demonstrated using a new dataset from the 2023 Kahramanmaraş earthquake in Türkiye, covering multiple cities with diverse urban settings. Results highlight that incorporating geospatial features significantly enhances detection performance and generalizability to previously unseen areas. By combining SAR imagery with detailed vulnerability and exposure information, our approach provides reliable and rapid building damage assessments without the dependency from available pre-event data. Moreover, the automated and scalable data generation process ensures the framework's applicability across diverse disaster-affected regions, underscoring its potential to support effective disaster management and recovery efforts. \textcolor{red}{Code and data will be made available upon acceptance of the paper.}
\end{abstract}


\IEEEpeerreviewmaketitle

\section{Introduction}
\label{intro}

\IEEEPARstart{B}{uilding} damage identification shortly after a disaster is critical for guiding emergency response and recovery efforts. Remote Sensing (RS) has long been used for this task \cite{FDAPG2012}, \cite{DONG201385}, \cite{Bruzzone2010}, initially via manual mapping and visual interpretation or simple change detection techniques (e.g. image differencing and ratioing of pre- and post-event imagery). However, these traditional approaches are labor intensive and sensitive to co-registration and lighting differences.

Although damage identification is often the first step, it is part of the broader process of building damage assessment (BDA), which includes not only detection but also the classification and quantification of damage severity.

Over the past decade, advances in using multi-sensor RS for seismic building vulnerability assessment have demonstrated that integrating Machine Learning (ML) approaches with multispectral and 3D structural information derived from satellite imagery can effectively characterize building vulnerabilities even before a disaster strikes \cite{Geiss2014}.

In recent years, the rise of Deep Learning (DL) has enabled more accurate and automated post-disaster BDA. Convolutional Neural Networks (CNNs) trained on large-scale datasets can learn complex visual patterns of damage, outperforming traditional visual assessment methods. For example, Ji \textit{et al.} \cite{Ji2019} showed that CNN-based features detected collapsed buildings with significantly higher accuracy than gray-level texture metrics in satellite images of past earthquakes. This success has spurred extensive research into DL models for BDA, aiming to localize buildings and classify their damage condition directly from aerial or satellite imagery.

A major catalyst behind the adoption of DL for BDA was the creation of large benchmark datasets and public challenges around 2019–2020. Notably, the xView2 challenge and its accompanying xBD dataset \cite{Gupta2019} provided the first massive corpus of annotated pre- and post-disaster images for building damage classification. This dataset contains over 850,000 building annotations across multiple disaster events (earthquakes, hurricanes, wildfires, etc.), with each building labeled by damage level (e.g. no damage, minor, major, or destroyed). This dataset and competition set a standard for the community, enabling researchers to train deep models that learn generic damage features across diverse scenarios. \href{https://github.com/DIUx-xView/xView2_first_place}{Winning solutions of xView2} demonstrated the effectiveness of modern CNN architectures in segmenting building footprints and predicting damage grades. Since then, numerous works have built upon these foundations. 

For instance, Valentijn \textit{et al.} \cite{Valentijn2020} showed that a CNN trained on xBD generalized well to new hazard events, especially when fine-tuned, indicating the model’s transferability across disasters. Kaur \textit{et al.} introduced a hierarchical transformer-based model for BDA on the xBD benchmark, achieving competitive F1 scores comparable to state-of-the-art CNN approaches and demonstrating the promise of transformer-based models in this domain \cite{Kaur2023}.

In essence, the xView2/xBD benchmark has become a springboard for BDA models, and it established the paradigm of using bi-temporal (pre- and post-event) high-resolution imagery as input for supervised DL techniques in damage mapping.

The devastating twin earthquakes of 6 February 2023 in Kahramanmaraş, Turkey (and neighboring regions in Türkiye and Syria) provided a real-world test bench for these advances. This event caused unprecedented building destruction over a vast area, and accordingly there have been numerous humanitarian initiatives and studies leveraging DL-based methods on very high resolution (VHR) optical imagery. A common strategy was to leverage models pre-trained on prior disaster data (such as xBD) and fine-tune them on the Turkey/Syria event to improve local performance.

For instance, Yılmaz \textit{et al.} \cite{Yilmaz2023} developed a radiometric change–analysis framework leveraging pre- and post-event Sentinel-1 SAR and Sentinel-2 MSI imagery to detect damaged buildings after the Kahramanmaraş earthquakes, reporting overall accuracies between approximately $74\%$ and $84\%$ across three test sites. Their study confirmed that transfer learning from a large prior dataset can yield highly accurate results even with relatively limited new training data. Karlberg \textit{et. al} \cite{Karlberg2023} implemented a dual-task U-Net trained on xBD and applied it to pre- and post-event VHR optical imagery of the February 6, 2023 Turkey earthquakes in Kahramanmaraş and Antakya. Their deep model achieved high F1 scores for undamaged buildings in Kahramanmaraş and Antakya, respectively, while the classification of destroyed buildings remained challenging. They found that, although DL and object-based methods performed comparably overall, DL still struggles with fine-grained damage grading when Ground Truth (GT) labels are scarce. 

Other teams integrated open-source data to facilitate fine-tuning: for instance, OpenStreetMap (OSM) building footprints and the Copernicus Emergency Management Service (EMS) post-event damage maps were combined to create training sets for Kahramanmaraş, against which pre-trained models (including the xView2 baseline) were adapted \cite{Kath2023}. These efforts, often deploying ensembles of CNN models \cite{Soleimani2023}, managed to produce damage maps at building scale covering the large affected area in a relatively short time.

Overall, the 2023 Turkey–Syria earthquake response demonstrated how state-of-the-art DL models, pre-trained on global disaster data and then fine-tuned to local conditions, can provide rapid and accurate damage assessments using optical satellite imagery with sub-meter spatial resolution.

Despite these promising developments, recent studies \cite{Wiguna2024} have shown that many DL-based approaches underperform when deployed in real emergency settings, where annotated data are unavailable and models must generalize to unseen disasters. This highlights the need for robust, data-efficient frameworks that can function reliably under operational constraints.

While most BDA research to date has focused on optical imagery, there is a growing interest in leveraging Synthetic Aperture Radar (SAR) for damage assessment. SAR sensors actively emit microwaves and capture the reflected signals, offering the main advantage of all-weather, day-or-night imaging. This makes SAR extremely valuable in operational scenarios; for example, immediately after an earthquake, optical coverage might be hindered by cloud cover or smoke, whereas radar can penetrate clouds and image the area regardless of lighting. Moreover, SAR backscatter and interferometric coherence are sensitive to structural changes: a building that collapses will alter the radar return (often causing decreased backscatter intensity and coherence loss) compared to its pre-disaster condition. These characteristics have been used in classical change detection approaches to infer damage “hotspots” when optical data were unavailable \cite{Wang2023}. For this purpose, in \cite{Voelker2024} two SAR-based damage proxy maps (DPMs) were utilized: a 30m C-band DPM by NASA’s Jet Propulsion Laboratory Advanced Rapid Imaging and Analysis (ARIA) team via coherence differencing of Sentinel-1 imagery, and a $\sim$7m L-band DPM by Argentina’s Comisión Nacional de Actividades Espaciales (CONAE) using coherence changes from SAOCOM-1 acquisitions.

Beyond these unsupervised coherence‐differencing products, researchers are now exploring DL models to exploit rich information of SAR for BDA. However, using SAR with DL comes with unique challenges: radar images suffer from speckle noise and geometric distortions (layover/shadowing), and the appearance of damage in SAR is less direct than in optical imagery (e.g. rubble and cracks are not immediately visible, but rather inferred from signal changes). Additionally, there have historically been fewer labeled SAR disaster datasets, making it harder to train DL models from scratch.

Despite these obstacles, initial progress is being made. Zhu \textit{et al.} \cite{ZhuBamler2021} review how DL can meet SAR, noting that specialized network designs and data preprocessing are needed to handle SAR’s peculiarities (e.g. complex-valued data, speckle statistics). On the application side, some studies have implemented bi-modal networks that fuse optical and SAR inputs to improve damage detection robustness. Jiang \textit{et al.} \cite{Jiang2020} proposed a deep homogeneous feature fusion (DHFF) model that uses a Siamese CNN to map heterogeneous optical and SAR image pairs into a common feature space, enabling change detection between pre- and post-event data even when one is optical and the other is SAR. This kind of cross-modal approach showed that combining SAR with optical can boost detection of damaged buildings that might be missed by optical alone.

Importantly, fully SAR-based DL methods for BDA are also emerging, driven by scenarios where optical data are unavailable or delayed. For example, Pang \textit{et al.} \cite{Pang2022} proposed CD-TransUNet, a hybrid Transformer–UNet that ingests pre- and post-event L-band SAR 'differential' images and employs coordinate attention and depthwise separable convolutions to segment building changes. In particular, Dietrich \textit{et al.} \cite{Dietrich2024} introduced an open source Google Earth Engine–based framework that leverages paired Sentinel-1 C-band time series and a Random Forest classifier to generate pixel-wise destruction probability heatmaps, which are then intersected with building footprints to produce nationwide, building-level damage assessments for Ukraine. These results demonstrated that deep networks can indeed identify damaged building footprints from SAR intensity images, although performance still lagged behind optical-based methods, partly due to the medium resolution of Sentinel-1, which hinders the detection of damage at the building scale, and the intrinsic noise of SAR data.

Other researchers have investigated polarimetric SAR (PolSAR) for damage assessment, leveraging the additional polarization information: Park and Jung \cite{Park2020} demonstrated that by analyzing changes in scattering parameters, such as Freeman–Durden decomposition powers, scattering matrix dissimilarity, and entropy, from pre- and post-earthquake PALSAR-2 PolSAR data, collapsed buildings could be detected with a $90.9\%$ detection rate and a $1.3\%$ false-alarm rate. Kim \textit{et al.} \cite{Kim2023} extended this approach by applying a contextual change analysis using GLCM-based textural features on bi-temporal Kompsat-5 X-band SAR data acquired in different polarization modes, achieving a $72.5\%$ detection rate and a $6.8\%$ false-alarm rate.  

A recurring theme in these SAR-DL studies is the data scarcity problem, since pre-disaster SAR images at very high resolution (VHR) are rarely collected systematically, training a model to recognize building damage in SAR often requires creative strategies such as simulation or transfer learning. Some studies have generated synthetic SAR imagery from optical data to enrich training sets by leveraging cross-domain attention and dual-conditional GANs such as CDA-GAN \cite{Wu2024CDA-GAN} and TSGAN \cite{Rangzan2023TSGAN} to produce realistic SAR samples. Others employ domain adaptation techniques to bridge the optical–SAR gap, such as Shi et al. \cite{Shi2022UDA} who developed a progressive-transfer UDA framework for ship detection by transferring knowledge from labeled optical datasets to unlabeled SAR images.

Despite relatively few publications so far, the use of DL on SAR for BDA is expected to grow, especially as new datasets become available. For example, the recent BRIGHT dataset \cite{BRIGHT2024} includes globally distributed disaster sites with both VHR optical and SAR imagery (from commercial SAR constellations like Capella and ICEYE) and detailed building damage labels. Such multimodal datasets will facilitate training models that can operate with either or both data sources, making damage mapping more resilient to data availability constraints.

Given this state-of-the-art, our work targets a significant remaining gap: enabling operational building damage detection using only post-event VHR SAR imagery, without relying on pre-disaster acquisitions. In many disaster scenarios, one cannot assume that a high-resolution SAR image of the affected city was acquired shortly before the event, unlike optical satellites which often have extensive archives. Even when pre-event VHR SAR images are available, they are usually acquired long before the disaster, introducing the risk of detecting unrelated changes that are not due to the event itself.

To address this constraint, we investigate how the integration of auxiliary geospatial data can enhance the reliability of post-event SAR-based detection, even in the absence of pre-event imagery. In particular, we evaluate how structural and exposure-related attributes can provide critical context to support the interpretation of SAR backscatter anomalies.

This approach is applied to the 2023 Kahramanmaraş earthquake in Türkiye, leveraging a multi-source data acquisition strategy that combines post-event COSMO-SkyMed (CSK) SAR imagery with complementary datasets, including:

\begin{itemize}
\item the Global Earthquake Model (GEM) exposure model, offering detailed building stock information (e.g., residential, commercial, industrial structures) derived from national and regional datasets for global earthquake risk assessment \cite{GEM2018Exposure};
\item OSM building footprints and post-event damage annotations for multiple affected Turkish cities, sourced from the \href{https://data.humdata.org/organization/hot}{Humanitarian Data Exchange (HDX)};
\item high-resolution Digital Surface Model (DSM) data from CartoSat-1 Euro-Maps, made available by ESA as part of Third Party Mission (TPM) data, serving as an auxiliary 3D reference for building damage assessment.
\end{itemize}

Similarly, Sun et al.~\cite{Sun2024} introduced the \href{https://github.com/ya0-sun/PostEQ-SARopt-BuildingDamage}{QuickQuakeBuildings dataset}, comprising over 4000 damage-annotated buildings from the 2023 Türkiye–Syria earthquake, co-registered with post-event VHR SAR and optical imagery. While their benchmark demonstrated that SAR-based damage detection is feasible, performance was consistently lower when using SAR inputs compared to optical ones, despite employing the same model architecture. This highlights the challenges associated with extracting reliable damage features from SAR imagery alone. However, the dataset itself was limited to a single urban area (Islahiye) and did not incorporate additional geospatial layers beyond building footprints.


Taking into account the lessons learnt from this previous experience, our study introduces a SAR-based post-event framework tailored for operational scalability and timely response in post-earthquake scenarios.

In contrast to existing datasets, our framework incorporates complementary structural and contextual information derived from high-resolution CartoSat DSM data and GEM exposure attributes. This supplementary information has been shown to improve damage detection by providing key structural and environmental features. The dataset covers multiple severely affected cities, including Türkoğlu, Islahiye, Osmaniye, Kahramanmaraş, and Nurdağı, enabling cross-city generalization and robust validation across diverse and previously unseen conditions.

Following \cite{Sun2024}, we formulate the damage detection task as an anomaly detection problem aimed at identifying buildings that exhibit structural changes in the immediate aftermath of an earthquake. While this task falls within the broader scope of BDA, our focus is specifically on binary classification, that is, determining whether a building is damaged or intact.

Integrating vulnerability and exposure information into the DL pipeline represents a key innovation of our approach. Most existing DL-based methods rely solely on image data, despite the well-established influence of factors such as building height, construction type, and local hazard intensity on damage assessment outcomes.

Specifically, building height (and related attributes like number of floors) acts as a proxy for structural vulnerability, while neighborhood-scale exposure attributes regarding building composition, construction materials, occupancy patterns, and population distribution further refine damage predictions. By integrating these auxiliary inputs directly into the DL pipeline, the model learns to correlate SAR-observed damage with underlying vulnerability and exposure factors.

To the best of our knowledge, this is the first time that fine-grained vulnerability and exposure attributes have been embedded inside a DL architecture for damage detection using VHR SAR imagery.

Recently, Rao et al. \cite{Rao2023} demonstrated the efficacy of such integrated methods, by fusing SAR-derived damage proxy maps, USGS ShakeMap intensities, and detailed building inventories within an ensemble ML framework, significantly surpassing SAR-only methods.

Our focus on binary damage detection reflects a practical need: in the immediate aftermath of a disaster, the first priority is to rapidly identify which structures are affected, so that emergency resources can be allocated quickly and effectively. Binary classification is also more feasible when GT data are limited or noisy, and it offers a more robust starting point for SAR-based analysis, where distinguishing between degrees of damage is often particularly challenging.

In summary, this work advances the field of post-disaster building damage detection by addressing a key operational gap: identifying damaged buildings using only post-event SAR imagery, without the dependency from available pre-disaster acquisitions.

Our main contributions are as follows:
\begin{itemize}
    \item we demonstrate the feasibility of accurate damage detection using only single-date VHR SAR imagery, which is often the only data available in the immediate aftermath of a disaster;
    \item we extend traditional damage detection methods into a more comprehensive building damage assessment framework by integrating additional data sources, specifically high-resolution DSM and GEM exposure attributes, that provide structural and socioeconomic context beyond commonly used inputs such as SAR imagery and building footprints;
    \item we show that our multimodal approach achieves accurate and consistent results across urban areas in a cross-city validation setup, generalizing well to unseen regions and even identifying damage missed by existing reference data; this suggests its potential to enhance broader damage mapping efforts and support ongoing assessment workflows;
    \item to this end, we construct a new dataset spanning several urban areas affected by the 2023 Kahramanmaraş earthquake and use it to assess our method under realistic, cross-city deployment conditions; the data pipeline is designed to be easily extensible to other disaster-affected regions, enabling future applications and comparative benchmarking.
\end{itemize}

The rest of this article is organized as follows. Section \ref{sec:study_area_data}  introduces the study area and describes the SAR, reference and auxiliary data used, along with the dataset creation procedure. Section~\ref{sec:methodology} details the proposed multimodal damage detection framework, including the data fusion strategy and network architecture. Section~\ref{sec:results} presents the experimental setup, evaluation metrics, and results across both cross-validation and unseen deployment scenarios. Section \ref{sec:conclusions} summarizes the main findings and outlines directions for future research.

\begin{figure*} [htp]
	\centering
	\adjustbox{max width=0.8\textwidth}{\includegraphics[]{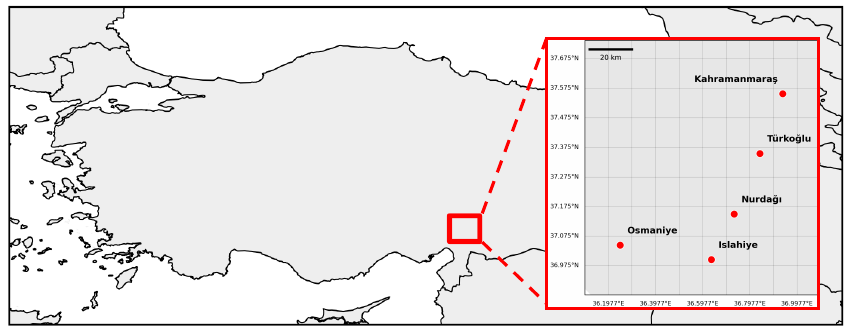}}
	\caption{Map of the study area showing the location of the Kahramanmaraş event supersite in southern Türkiye, with a zoomed-in view of the spatial distribution of the five cities analyzed in this study: Islahiye, Osmaniye, Nurdağı, Türkoğlu, and Kahramanmaraş.} \label{fig:study_area}
\end{figure*}

\section{Data and Study Area}\label{sec:study_area_data}
On 6 February 2023 a magnitude 7.8 earthquake struck near Kahramanmaraş in southeastern Türkiye, followed nine hours later by a magnitude 7.5 aftershock near Elbistan. The sequence occurred along the East and North Anatolian faults, which delineate the boundary between the Anatolian, Eurasian and Arabian tectonic plates. Both events comprised shallow ruptures ($\sim18$km and $\sim10$km depth, respectively) along different segments of the East Anatolian Fault, together rupturing approximately $500$–$600$ km of fault  (\href{https://www.ingv.it/en/resources-and-services/environment-earthquakes-and-volcanoes/reports-reports-and-reports/seismic-sequence-in-south-eastern-T%C3%BCrkiye}{Istituto Nazionale di Geofisica e Vulcanologia (INGV)}). As the strongest event in the region since $1939$, it affected eleven provinces over roughly $350,000$ km$^2$, claimed over $53,000$ lives in Türkiye and more than $8,000$ in Syria, and left about 1.5 million people homeless. Direct economic losses are estimated at USD $34.2$ billion, nearly $4\%$ of Türkiye’s 2021 GDP \href{https://www.worldbank.org/en/news/press-release/2023/02/27/earthquake-damage-in-turkiye-estimated-to-exceed-34-billion-world-bank-disaster-assessment-report}{World Bank}.

\begin{table}[hbt]
  \centering
  \caption{COSMO-SkyMed SAR post-event data used for this study}
  \label{tab:csk_data}
   \resizebox{0.95\columnwidth}{!}{%
  \begin{tabular}{l l l l}
    \toprule
    \textbf{City}          & \textbf{Acquisition date} & \textbf{Acquisition mode}  & \textbf{Polarization} \\
    \midrule
    Islahiye               & 2023-02-25                & Stripmap HIMAGE            & HH                   \\
    Kahramanmaraş          & 2023-03-01                & Stripmap HIMAGE            & HH                   \\
    Nurdağı                & 2023-02-25                & Stripmap HIMAGE            & HH                   \\
    Osmaniye               & 2023-02-15                & Stripmap HIMAGE            & HH                   \\
    Türkoğlu               & 2023-02-25                & Stripmap HIMAGE            & HH                   \\
    \bottomrule
  \end{tabular}
  }
\end{table}

In this study, we examine five cities belonging to this event supersite, each illustrating different aspects of seismic vulnerability and exposure. Their location and spatial distribution are shown in Figure \ref{fig:study_area}. Kahramanmaraş, with a population of $571,266$ residents in its urban core, experienced the heaviest damage concentrated within its densely built, mixed-use neighborhoods. Further west, on the slopes of the Nur Mountains, the town of Islahiye suffered extensive structural damage due to foundation failures and widespread landslides. To the north, the agricultural district of Nurdağı saw near-complete destruction of its low-rise masonry buildings. To the south, Osmaniye, with approximately $244,500$ residents and serving as an important transportation hub, displays a clear contrast in damage patterns between older concrete frame structures built before $1980$ and newer reinforced concrete buildings. Lastly, Türkoglu, which has experienced rapid population growth in recent decades, has a significant number of unreinforced brick buildings, which proved particularly vulnerable to collapse.

\subsection{Input Data}
The SAR data utilized in this study is sourced from the COSMO-SkyMed (CSK) satellite constellation operated by the Italian Space Agency (ASI). This satellite system is particularly suited for disaster management due to its all-weather, day-or-night imaging capabilities provided by active microwave emission and its high spatial resolution ($2.5$m) and very short time response \cite{CSKRef1}, \cite{ASI2016}. Specifically, our approach is tested using SAR imagery made available to support institutions, emergency management services, and within the framework of international cooperation for disaster mapping and response.

This essential condition was made possible due to the availability of a comprehensive SAR imagery archive regularly collected as part of ASI’s CSK background mission, a low-priority acquisition plan designed to maximize system exploitation and consistently accumulate strategic historical datasets. In this specific case, ASI enhanced this collection by acquiring additional imagery within the framework of the Committee on Earth Observation Satellites (CEOS), explicitly supporting the GEO Geohazard Supersites and Natural Laboratory (GSNL) initiative focused on the \href{https://geo-gsnl.org/supersites/event-supersites/active-event-supersites/kahramanmaras-event-supersite/}{Kahramanmaraş Event Supersite} and its dedicated \href{https://geo-gsnl.org/supersites/event-supersites/active-event-supersites/kahramanmaras-event-supersite/eo-data-access-for-the-kahramanmaras-event-supersite/}{EO data access portal}. 
\newline Technical specifications of the acquired CSK data are detailed in Table~\ref{tab:csk_data}, which summarizes the acquisition dates, acquisition mode, and polarization for satellite images covering different urban sites. For consistency across all five cities, every post-event image was acquired with the same satellite sensor (\textit{CSK4}) and the identical acquisition mode (\textit{Stripmap HIMAGE}) in HH polarization, thereby minimizing, as far as possible, variations in image geometry and system noise that could affect damage signature extraction across different urban environments. 

\begin{figure*} [htp]
	\centering
	\adjustbox{max width=\textwidth}{\includegraphics[]{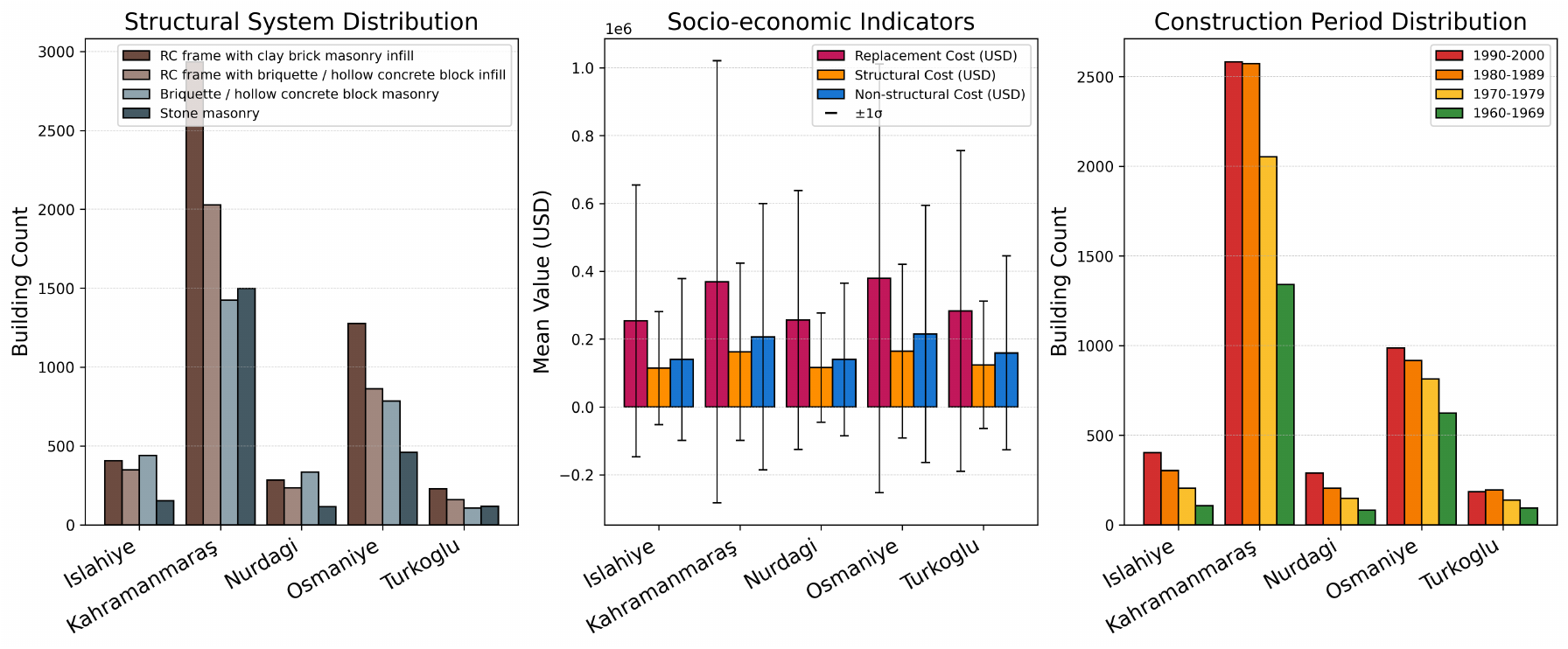}}
	\caption{Summary of GEM exposure metrics for the five analyzed cities. The left panel shows the distribution of structural typologies, such as reinforced concrete (RC) frames with clay brick or concrete-block infill, briquette/block masonry, and stone masonry, highlighting construction practices that affect seismic vulnerability. The central panel reports city-level means and standard deviations for key socioeconomic indicators from the GEM dataset, including replacement cost, structural cost, and non-structural cost, which reflect potential reconstruction value and population exposure. The right panel illustrates the age distribution of the building stock by construction period (1960–1969, 1970–1979, 1980–1989, 1990–2000), used here as a proxy for seismic fragility. 
    } \label{fig:gem_statistics}
\end{figure*}

\subsection{Reference and Auxiliary Data}
In addition to SAR data, we integrated complementary geospatial datasets to enhance structural context and provide reliable damage references. Specifically, we incorporated building footprints from the \href{https://data.humdata.org/dataset/hotosm_tur_buildings}{HOTOSM Turkey Buildings} layer and validated post-event polygons indicating collapsed buildings from the \href{https://data.humdata.org/dataset/hotosm_tur_destroyed_buildings}{HOTOSM Turkey Destroyed Buildings} layer, both sourced from OpenStreetMap (OSM). 
It is important to acknowledge, however, that the OSM annotations used as reference data in this study introduce a degree of label uncertainty, given the rapid and volunteer-based nature of post-disaster mapping. To address this, we manually reviewed ambiguous cases using high-resolution Google Earth imagery, and in several instances the model correctly identified damage that was missing from the OSM annotations, indicating the likely presence of false negatives, as later illustrated in Figure~\ref{fig:patch_comparisons}. 
\newline For elevation and structural context, we utilized Digital Surface Model (DSM) data from the CartoSat-1 Euro-Maps product, accessed under ESA’s Third-Party Mission (TPM) programme. These DSMs, generated from systematically acquired CartoSat-1 stereo imagery, offer a 5-meter spatial resolution and vertical accuracies ranging from 5 to 10 meters, suitable for capturing building roof heights and surrounding terrain features. The Euro-Maps DSMs (Level A and A+) include bare-surface height and optional orthorectified imagery layers, providing both geometric and radiometric references for vulnerability modeling. Access to the CartoSat-1 DSM archive was facilitated by GAF AG through an ESA-GAF agreement under the TPM programme, which continues to grant free-of-charge data access for scientific users upon submission of project proposals (\href{https://earth.esa.int/eogateway/catalog/cartosat-1-archive-and-euro-maps-3d-digital-surface-model}{ESA TPM Catalogue}). Access to these data was granted through the project ID PP0101796.
\newline 
It is also important to note that the DSM refers to the pre-event context, as it was generated several years prior to the earthquake, and is thus used primarily to characterize the height profiles of buildings and surrounding terrain.
\newline
Additionally, to incorporate socioeconomic and structural vulnerability characteristics, we utilized building exposure data provided by the Global Exposure Model (GEM). The high-resolution GEM exposure dataset was obtained through a formal request on the GEM data portal and completion of the corresponding license agreement with the GEM Foundation. This dataset includes critical feature attributes at the neighborhood scale, such as structural typology (e.g., reinforced concrete versus masonry), replacement cost values (representing economic exposure in light of post-event adjustments), estimated average occupancy per asset, and the decade of construction of the building stock.
Although building age is not a direct measure of seismic vulnerability, it can serve as a practical proxy. The underlying assumption is that older buildings are, on average, less likely to meet current seismic design standards. Of course, this is not always the case: some older structures may actually perform better than more recent ones due to robust traditional construction methods, while even relatively new buildings (e.g., from the 1990s) may prove vulnerable if poorly designed or built. Still, in the absence of detailed structural information, age-based stratification remains a widely accepted and practical approach for estimating differences in vulnerability at the city scale \cite{Lagomarsino2006}.

The statistics of some of these features for different cities are summarized in Figure~\ref{fig:gem_statistics}.

\begin{table}[hbt]
    \centering
    \caption{Distribution of intact versus damaged buildings across selected cities following dataset generation.}
        \begin{tabular}{l rr}
            \toprule
            \textbf{City}        & \textbf{\#Intact} & \textbf{\#Damaged} \\
            \midrule
            Islahiye             & 3{,}825           & 192                  \\
            Kahramanmaraş        & 4{,}641           & 233                  \\
            Nurdağı              & 3{,}289           & 498                  \\
            Osmaniye             & 317               & 16                   \\
            Türkoğlu             & 453               & 23                   \\
            \bottomrule
        \end{tabular}
    \label{tab:building_counts}
\end{table}

\begin{figure}[ht]
  \centering
  \includegraphics[width=0.8\columnwidth]{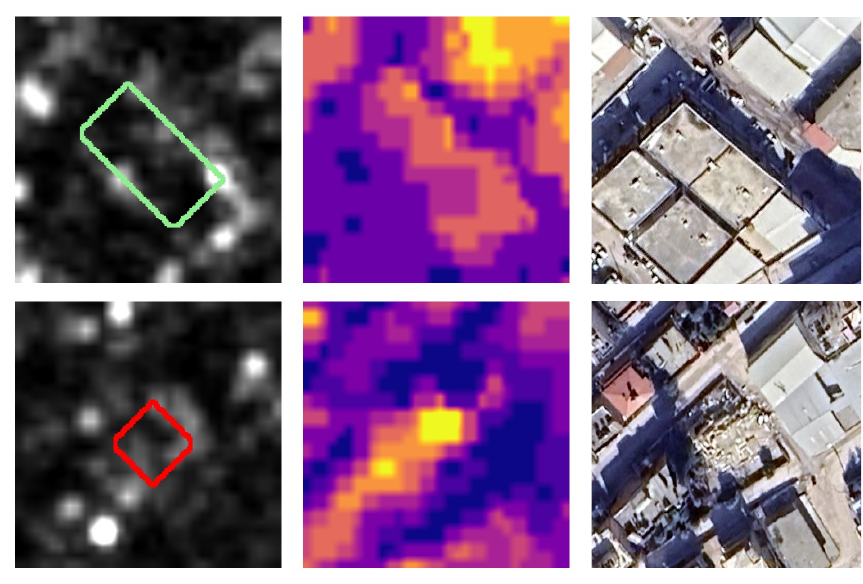}
  \caption{Examples of building-centered image patches from our dataset used for building damage detection. Each row corresponds to a single sample: the first one shows an intact building (negative class), while the second shows a collapsed building (positive class). From left to right, the panels display the post-event SAR backscatter patch with the building footprint outlined, the corresponding DSM patch, and a high-resolution optical image from Google Earth used as a comparative basemap for visual reference.}
\label{ref:sar_dsm_footprint_patches}
\end{figure}

\subsection{Dataset Creation}
For each city, post-event SAR imagery and high-resolution DSM data were collected and georeferenced to a common coordinate reference system (CRS). Corresponding OSM building footprints were also integrated to accurately identify anomalies in SAR backscatter indicative of structural damage or collapse. Building footprints are crucial in improving earthquake damage detection accuracy using SAR imagery by providing essential spatial context, and the increasing availability of up-to-date footprints from open platforms (OSM itself and local government datasets) makes their integration both practical and reliable. Previous studies have validated the efficacy of integrating building footprints into SAR \cite{Sun2020} and specifically SAR-based anomaly detection frameworks \cite{Sun2024}.  
\newline Finally, GEM exposure attributes were associated with each building by identifying the nearest GEM data point to each building centroid, ensuring that each structure was comprehensively enriched with relevant contextual information such as structural typology, construction period, replacement cost, and average occupancy.  
However, it is important to note that this association was necessary because the GEM dataset provides aggregated statistics for groups of buildings within each spatial reference cell. As a result, the assignment of these attributes to individual building centroids introduces a degree of uncertainty due to the limited spatial resolution and aggregation level of the original GEM data.
\newline To address computational challenges associated with processing extensive urban datasets, a selective sampling strategy was adopted. Specifically, all buildings labeled as destroyed were included, with intact buildings randomly sampled at a ratio of 20 intact buildings for every damaged one. This $1$:$20$ ratio reflects the natural imbalance commonly observed in earthquake scenarios, where damaged buildings represent anomalies, aligning well with anomaly detection methodologies.  
\newline This sampling strategy resulted in the creation of $13,487$ building-centered SAR patches, each paired with the corresponding DSM patch, building footprint mask, and normalized GEM feature vector. Their distribution into damaged (positive) and intact (negative) classes across the five cities is summarized in Table~\ref{tab:building_counts}.  
\newline For each selected building, a $32 \times 32$ pixel image patch centered on the building centroid was extracted from both SAR and DSM datasets. The patch size was chosen to isolate individual buildings with minimal surrounding context, thereby emphasizing localized damage signals while reducing background noise. Some examples of building centered SAR patches with their corresponding footprint outlines, DSM visualizations, and the matching high-resolution post-event optical imagery from Google Earth are shown in Figure~\ref{ref:sar_dsm_footprint_patches}.
\newline It should be noted that the dataset creation procedure is fully automated, enabling efficient and scalable extension to other areas similarly impacted by disasters, thus facilitating rapid and consistent damage assessment across various geographic regions.

\begin{figure*} [htp]
	\centering
	\adjustbox{max width=\textwidth}{\includegraphics[]{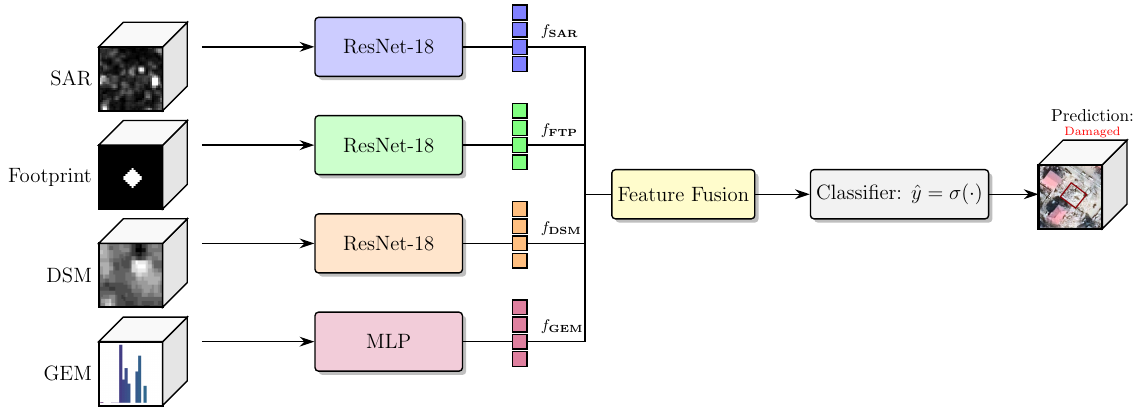}}
	\caption{Schematic of the proposed late-fusion multimodal network for building damage assessment. The architecture processes four input modalities: SAR image patches, building footprint masks (each centered on the building of interest), DSM patches, and GEM-derived feature vectors. Each modality is processed by a dedicated feature extractor, with ResNet-18 backbones used for SAR, footprint, and DSM inputs, and a multi-layer perceptron (MLP) applied to the GEM attributes. The resulting feature vectors are concatenated and passed to a classification head consisting of two fully connected layers with ReLU activations and dropout. The final output is a binary damage prediction.}\label{fig:workflow}
\end{figure*}

\section{Methodology}\label{sec:methodology}
We utilize a modular late fusion multimodal neural network for building damage detection. The model operates on post-disaster CSK imagery, enriched with auxiliary geospatial information. Building upon the baseline architecture proposed by Sun et al.~\cite{Sun2024}, which uses SAR imagery and building footprint masks, our approach introduces two additional data sources: DSM patches and exposure-related attributes from GEM. These enhancements aim to improve detection accuracy and generalization across different urban environments.

Figure~\ref{fig:workflow} illustrates the architecture, which processes four input modalities:
\begin{itemize}
\item $I_{\text{SAR}}$: a post-event SAR patch centered on the building;
\item $M_{\text{FTP}}$: a binary mask patch representing the building footprint;
\item $I_{\text{DSM}}$: a DSM patch providing structural context;
\item $\mathbf{x}_{\text{GEM}}$: a vector of GEM-derived vulnerability and exposure features.
\end{itemize}

Each spatial input ($I_{\text{SAR}}$, $M_{\text{FTP}}$, $I_{\text{DSM}}$) is independently processed by a ResNet-18 encoder, yielding modality-specific embeddings $\mathbf{f}_{\text{SAR}}$, $\mathbf{f}_{\text{FTP}}$, and $\mathbf{f}_{\text{DSM}}$, respectively:
\begin{equation}
\mathbf{f}_k = \text{ResNet18}(X_k), \quad X_k \in { I_{\text{SAR}}, M_{\text{FTP}}, I_{\text{DSM}} }
\end{equation}

The GEM attribute vector $\mathbf{x}_{\text{GEM}}$ is transformed into a dense feature embedding via a multi-layer perceptron (MLP), consisting of fully connected layers with non-linear activations:
\begin{equation}
\mathbf{f}_{\text{GEM}} = \text{MLP}(\mathbf{x}_{\text{GEM}})
\end{equation}

This choice reflects the nature of GEM features, which are tabular and encode attributes such as building typology, construction period, and occupancy at the neighborhood scale. Unlike image-based inputs, these data lack spatial structure and localized patterns, making convolutional encoders like ResNet unsuitable for feature extraction in this context.

All feature embeddings are then concatenated to form a unified multimodal representation:
\begin{equation}
\mathbf{f}_{\text{LF}} = [ \mathbf{f}_{\text{SAR}}, \mathbf{f}_{\text{FTP}}, \mathbf{f}_{\text{DSM}}, \mathbf{f}_{\text{GEM}} ]
\end{equation}

This late fusion feature vector $\mathbf{f}_{\text{LF}}$ is passed through a classification head composed of two fully connected layers with ReLU activations and dropout, with $\text{FC}_{\text{CLS}}(\cdot)$ denoting this classification module. The final output $\hat{y}$ is the predicted probability of building damage, obtained via sigmoid ($\sigma$) activation:
\begin{equation}
\hat{y} = \sigma\left( \text{FC}_{\text{CLS}}(\mathbf{f}_{\text{LF}}) \right)
\end{equation}
where $\sigma(\cdot)$ is the sigmoid function used for binary classification.

The model is trained end-to-end using the binary cross-entropy loss:
\begin{equation}
\mathcal{L}(\mathbf{y}, \hat{\mathbf{y}}) = -\frac{1}{N} \sum_{i=1}^{N} \left[ y_i \log(\hat{y}_i) + (1 - y_i) \log(1 - \hat{y}_i) \right]
\end{equation}
where $y_i \in \{0, 1\}$ is the GT damage label for the $i$-th building.

The modular design allows for flexible inclusion or omission of DSM and GEM inputs during both training and inference, making the model adaptable to varying data availability in real-world disaster response contexts.


\begin{figure*}[htp]
	\centering
	\includegraphics[width=0.9\textwidth]{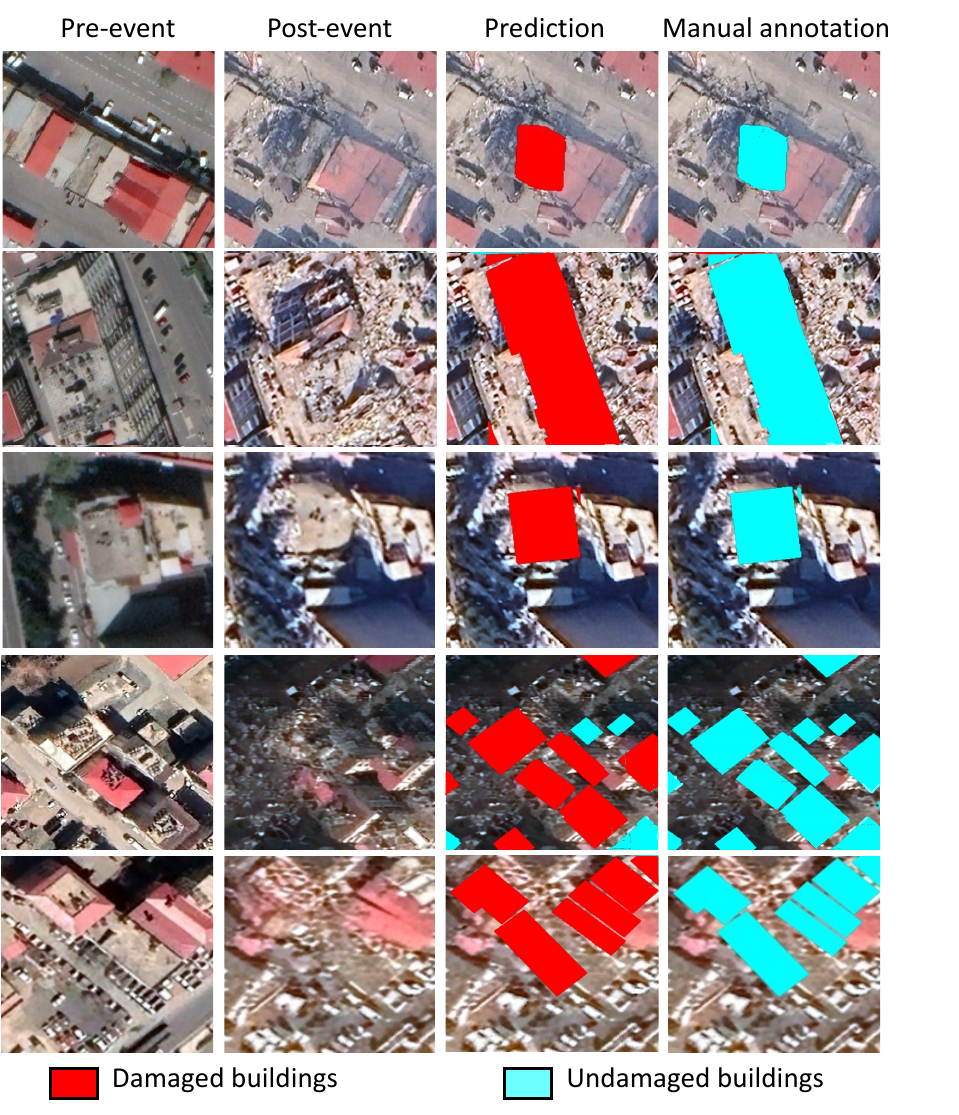}
        \caption{Qualitative examples of the model’s performance on cities excluded from training. Samples are drawn from Kahramanmaraş (first three rows) and Nurdağı (last two rows). Each row shows, from left to right: pre-event imagery, post-event imagery, the model’s damage prediction, and the manual annotations overlaid on a VHR Google Earth basemap matched to the respective pre- and post-event acquisition dates. These examples highlight cases in which the model successfully identifies damaged buildings that were mislabeled in the manual annotations, illustrating its potential to assist and refine manual damage assessments efforts.}
	\label{fig:patch_comparisons}
\end{figure*}

\section{Experiments and Damage Detection Results}\label{sec:results}
In this section, we present and analyze the outcomes of our experimental evaluation, structured into two main parts: in-distribution and out-of-distribution assessments. These experiments validate the effectiveness of the proposed multimodal architecture and emphasize its robustness and operational readiness in realistic disaster-response scenarios. Crucially, our out-of-distribution experiments address the operational requirement of deploying models in urban environments completely unseen during training, thus simulating realistic scenarios encountered in emergency situations following disasters. All experiments were conducted using the same hyperparameter configuration, consistent with Sun et al.~\cite{Sun2024}, to ensure comparability and isolate the contribution of each additional data modality.

\subsection{Experimental Setup}
Three complementary setups were designed to evaluate the model performance under different operational scenarios:

\begin{enumerate}
    \item \textbf{Stratified cross-validation across all cities}: a 5-fold stratified cross-validation was performed on the entire dataset, including all buildings from the five cities. This configuration assesses the model's learning behavior and stability across diverse urban environments while maintaining geographic consistency between training and validation data.
    
    \item \textbf{Generalization to unseen cities}: a leave-one-city-out evaluation was performed to simulate operational deployment in disaster-affected areas not represented during training. This tests the model's robustness and adaptability to spatial domains with different structural and socioeconomic characteristics.

    \item \textbf{Comparative benchmarking}: a targeted evaluation on the Islahiye subset enables a direct comparison with the method proposed by Sun et al. \cite{Sun2024}. All experiments, including this one, adopt the same multimodal network architecture and training procedure, allowing a consistent and fair benchmark across methods.
\end{enumerate}

\begin{table*}[ht]
    \caption[
        Average performance across 5-fold cross-validation
    ]{
        \parbox{0.7\textwidth}{
        \centering
        \vspace{0.1cm}
        Average performance and standard deviation across 5-fold cross-validation for each input configuration. 
        All metrics refer to the positive class (damaged buildings). Precision, recall, F1 score, and Cohen’s kappa are reported at the threshold that maximizes the F1 score on the precision–recall curve, while AUROC is computed over all thresholds. Best results per metric are highlighted in bold.
        }
    }
    \centering
    \begin{tabular}{lccc}
        \toprule
        \textbf{Metric} & \textbf{\textit{SAR + FTP}} & \textbf{\textit{SAR + FTP + DSM}} & \textbf{\textit{SAR + FTP + DSM + GEM}} \\
        \midrule
        Precision & $0.914 \pm 0.034$ & $\textbf{0.939} \pm \textbf{0.029}$ & $0.922 \pm 0.017$ \\
        Recall    & $0.819 \pm 0.042$ & $0.835 \pm 0.030$ & $\textbf{0.852} \pm \textbf{0.013}$ \\
        F1‐Score  & $0.863 \pm 0.029$ & $0.883 \pm 0.015$ & $\textbf{0.886} \pm \textbf{0.007}$ \\
        AUROC     & $0.960 \pm 0.007$ & $0.967 \pm 0.004 $ & $\textbf{0.968} \pm \textbf{0.005}$ \\
        Kappa     & $0.849 \pm 0.031$ & $0.872 \pm 0.016$ & $\textbf{0.874} \pm \textbf{0.007}$ \\
        \bottomrule
    \end{tabular}
    \label{tab:fold_stats}
\end{table*}

\begin{table*}[ht]
    \caption{
        \parbox{0.8\textwidth}{
        \centering
        \vspace{0.1cm}
        Performance on each city when excluded from training, simulating emergency deployment in unseen urban areas. 
        The table presents results for all input configurations. 
        Best values per city are shown in bold. 
        All metrics are computed on the positive class (damaged buildings). 
        Precision, recall, F1 score, and Cohen’s kappa are reported at the threshold that maximizes the F1 score on the precision–recall curve; AUROC is computed across all thresholds.
        }
    }
    \centering
    \begin{tabular}{lccccc c}
        \toprule
        \textbf{City} & \textbf{Simulation} & \textbf{Precision} & \textbf{Recall} & \textbf{F1 score} & \textbf{AUROC} & \textbf{Kappa} \\
        \midrule
        Islahiye     & \textit{SAR + FTP}               &   0.127    &  \textbf{0.396}    &   0.192      &  0.683     &    0.129   \\
                     & \textit{SAR + FTP + DSM}         &  \textbf{0.312}   &   0.260           &  \textbf{0.284}      &  \textbf{0.750}    &  \textbf{0.252}   \\
                     & \textit{SAR + FTP + DSM + GEM}   &   0.242    &   0.307           &   0.271      &   0.720    &   0.229   \\
        \midrule
        Kahramanmaraş& \textit{SAR + FTP}               &   0.201    &  \textbf{0.476}    &   0.282      &  0.809     &   0.231    \\
                     & \textit{SAR + FTP + DSM}         &  \textbf{0.361}   &   0.339           &   0.350      &  0.824     &   0.318   \\
                     & \textit{SAR + FTP + DSM + GEM}   &   0.358    &   0.408           &  \textbf{0.382}      &  \textbf{0.853}    &  \textbf{0.348}    \\
        \midrule
        Nurdağı      & \textit{SAR + FTP}               &   0.240    &   0.240           &   0.318      &  0.661     &   0.174    \\
                     & \textit{SAR + FTP + DSM}         &   0.255    &  \textbf{0.470}    &   0.331      &  \textbf{0.669}    &   0.194   \\
                     & \textit{SAR + FTP + DSM + GEM}   &  \textbf{0.276}   &   0.416           &  \textbf{0.332}      &  0.668     &  \textbf{0.207}    \\
        \midrule
        Osmaniye     & \textit{SAR + FTP}               &   0.063    &  \textbf{0.812}    &   0.118      &  0.552     &   0.031   \\
                     & \textit{SAR + FTP + DSM}         &   0.106    &   0.106           &   0.143      &  0.606     &   0.106     \\
                     & \textit{SAR + FTP + DSM + GEM}   &  \textbf{0.150}   &   0.188           &  \textbf{0.167}      &  \textbf{0.625}    &  \textbf{0.120}    \\
        \midrule
        Türkoğlu     & \textit{SAR + FTP}               &   0.217    &   0.217           &   0.217      &  0.648     &   0.178     \\
                     & \textit{SAR + FTP + DSM}         &   0.226    &   0.304           &   0.259      &  0.631     &   0.216    \\
                     & \textit{SAR + FTP + DSM + GEM}   &  \textbf{0.333}   &  \textbf{0.348}    &  \textbf{0.340}      &  \textbf{0.653}    &  \textbf{0.306}     \\
        \bottomrule
    \end{tabular}
    \label{tab:metrics}
\end{table*}

\subsection{Performance Metrics}
Performance was quantified using precision, recall, F1 score, Cohen’s kappa, and AUROC, all computed with respect to the positive class (damaged buildings). Classification thresholds were selected based on the value that maximized the F1 score on the precision–recall curve, in order to account for class imbalance. AUROC was used as a threshold-independent measure of the model’s discriminative ability.

\subsection{Cross-Validation Among Multiple Cities}
\begin{figure}[htp]
    \centering
    \includegraphics[width=0.4\textwidth]{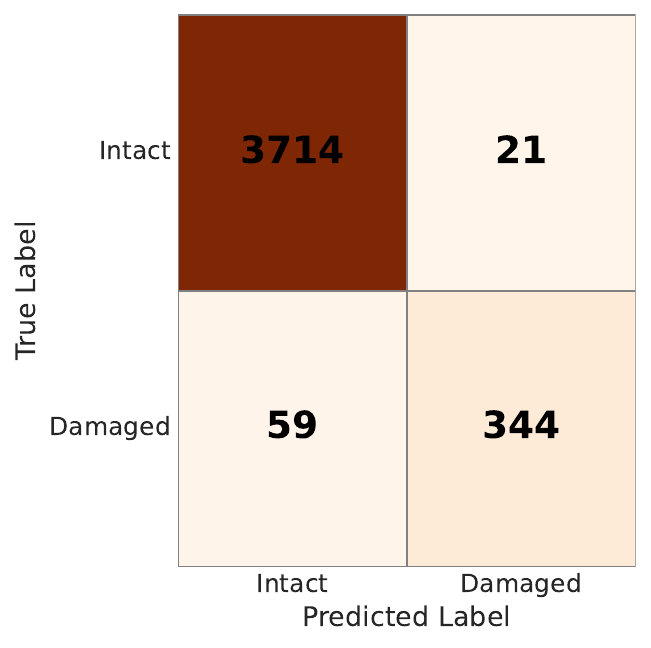}
    \caption{Confusion matrix for the cross-validation fold achieving the highest F1 score. Each cell shows the raw counts of true negatives (top-left), false positives (top-right), false negatives (bottom-left) and true positives (bottom-right) for the binary damage classification (Undamaged vs. Damaged).}
    \label{fig:impact_comparison}
\end{figure}

Table~\ref{tab:fold_stats} reports the average cross-validation performance across five folds, highlighting the incremental benefits of each additional data source. Starting from the baseline model using SAR and footprint data, the integration of DSM information consistently improved performance across all metrics, particularly in terms of Precision and F1 score. Adding GEM exposure attributes further enhanced the model, yielding the highest recall and the most stable classification behavior overall. Although performance differences between configurations are moderate, they are consistent and indicative of the added value offered by each additional data source. Figure \ref{fig:impact_comparison} shows the confusion matrix for the fold achieving the highest F1 score, highlighting the strong ability of the model to accurately detect damaged buildings while minimizing false positives and false negatives.

\subsection{Generalization to Unseen Cities}

Table~\ref{tab:metrics} reports model performance when each city is used as test area, while training is performed on the remaining four. This setup emulates realistic deployment scenarios where the model is applied to urban areas it has not been exposed to during training. As expected, performance varies across cities due to differences in building types, annotation quality, and exposure conditions. Nevertheless, the integration of DSM and GEM data leads to consistent and often meaningful improvements. DSM enhances the model’s ability to capture structural characteristics, as reflected by higher F1 scores and Kappa values. GEM further improves recall and model stability, especially in complex environments like Kahramanmaraş and Türkoğlu, where the full configuration achieves the strongest overall performance. These results demonstrate how incorporating structural and contextual information helps the model adapt more effectively to new geographical settings.

Complementing the quantitative results, qualitative analysis provides additional insight into model behavior. As illustrated in Figure~\ref{fig:patch_comparisons}, the network is able to correctly identify damaged buildings in cities not seen during training, even when these instances are mislabeled as intact in the GT. Several of these discrepancies were verified through inspection of high-resolution post-event imagery from Google Earth, revealing annotation errors in crowdsourced datasets such as OSM. These examples suggest that the model can also support the refinement of manual or volunteer-based damage annotations. In this light, the true reliability of the proposed multimodal framework may be higher than what quantitative metrics alone suggest.

\begin{table}[ht]
    \centering
    \caption{Comparative results (5-fold cross-validation) for the Islahiye area. Metrics computed at best-F1 threshold. Best results per metric in bold.}
    \resizebox{\columnwidth}{!}{%
    \begin{tabular}{lccc}
        \toprule
        \textbf{Metric} & \textbf{Sun et al. \cite{Sun2024}$^*$} & \textbf{This paper (\textit{SAR + FTP})} & \textbf{\textbf{This paper (All inputs)}$^{**}$} \\
        \midrule
        Precision & $0.237 \pm 0.116$ & $0.203 \pm 0.045$ & $\mathbf{0.255} \pm \mathbf{0.070}$ \\
        Recall & $0.427 \pm 0.104$ & $\mathbf{0.433} \pm \mathbf{0.093}$ & $0.395 \pm 0.099$ \\
        F1 score & $0.282 \pm 0.077$ & $0.269 \pm 0.037$ & $\mathbf{0.295} \pm \mathbf{0.038}$ \\
        AUROC & $\mathbf{0.769} \pm \mathbf{0.032}$ & $0.769 \pm 0.038$ & $0.763 \pm 0.037$ \\
        Kappa & -- & $0.219 \pm 0.043$ & $\mathbf{0.251} \pm \mathbf{0.044}$ \\
        \bottomrule
    \end{tabular}
    }
\vspace{-3mm}
    \begin{minipage}{\columnwidth}
        \centering
        \tiny{\textit{$^*$ SAR data from Capella Space.\\ 
        $^{**}$ It refers to the combination of SAR, footprint, DSM and GEM.}}
    \end{minipage}
\label{tab:comparison_sun}
\end{table}

\subsection{Comparative Analysis with Previous Work}
To further assess the effectiveness of our approach, we compared our results with the recent study by Sun et al. \cite{Sun2024}, which was focused on the Islahiye area.  
For this purpose, we applied the 5-fold cross-validation procedure described earlier, restricted to the Islahiye subset.

Table~\ref{tab:comparison_sun} summarizes the results. Our baseline configuration using SAR and footprint data achieves performance that is overall comparable to that reported by Sun et al., despite relying on SAR imagery from a different sensor (COSMO-SkyMed instead of Capella Space). Although their model shows slightly better results in some metrics, particularly precision and F1 score, this difference may be partly explained by the higher spatial resolution of Capella imagery (approximately 0.5 m) compared to the 2.5 m resolution of the COSMO-SkyMed StripMap mode used in our study. Still, the overall alignment in performance supports the consistency of our dataset and confirms that the experimental conditions are methodologically comparable and reproducible.

When DSM and GEM data are included, our enhanced multimodal model achieves the best results across key metrics, particularly F1 score and Cohen’s Kappa. These improvements reflect a better balance between precision and recall, and increased reliability in classification. The added value of topographic and exposure information underscores the importance of integrating structural and contextual features for accurate damage detection.

\section{Conclusions} \label{sec:conclusions}
This work proposed a multimodal DL framework for post-earthquake building damage assessment using VHR SAR imagery from the COSMO-SkyMed constellation, enriched with auxiliary geospatial data including DSM and GEM exposure attributes. Unlike traditional bi-temporal or optical-based approaches, our method operates using only post-event data, addressing critical constraints in real-world emergency scenarios where timely-collected or cloud-free imagery (either before or after the event) may be unavailable.

Results from both cross-validation and cross-city evaluation on five urban areas affected by the 2023 Kahramanmaraş earthquake demonstrate the benefit of incorporating structural and contextual information. The integration of DSM and GEM features led to improved F1 score, recall, and more consistent performance, particularly when applied to cities not seen during training. Qualitative assessments further confirmed the model’s ability to identify damage not captured by reference annotations, underscoring its potential to assist or refine manual mapping efforts.

Future research will focus on the adoption of more advanced algorithmic techniques, such as attention-based mechanisms and hybrid CNN-transformer architectures, to enhance performance and interpretability. Furthermore, we aim to expand the training dataset with additional urban areas and hazard contexts to improve generalization. Future work will also attempt to model the uncertainty associated with the crowdsourced damage annotations and spatially aggregated exposure data used in this study.

\section*{Acknowledgments}
This research was performed in the framework of a PhD funded by NextGenerationEU, Action 4, DM n. 118, 02/03/2023. COSMO-SkyMed Products © of the Italian Space Agency (ASI) were accessed under a license to use granted by ASI within the Project Card TurkeyEQ2024. Access to CartoSat-1 Euro-Maps 3D data was granted through ESA’s Third-Party Mission (TPM) programme under Project-ID PP0101796 (“Turkey Earthquake 3D Data”), with data delivery facilitated by GAF AG. The Detailed Building Exposure Model of Turkey was provided by the GEM Foundation as part of its efforts toward global seismic risk assessment, and was accessed under a license agreement for academic research use.

\bibliographystyle{IEEEtran.bst}
\bibliography{references}

\begin{minipage}[t]{0.47\textwidth}
\begin{IEEEbiography}[{\includegraphics[width=1in,height=1.25in,clip,keepaspectratio]{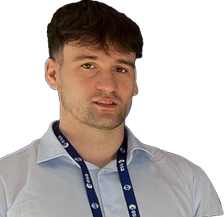}}]{Luigi Russo}
Student Member, IEEE, earned his master's degree (cum laude) in Electronic Engineering for Automation and Telecommunications from the University of Sannio, Benevento, Italy, in 2023. He is currently working toward the Ph.D. degree in Electronics, Computer Science and Electrical Engineering at the University of Pavia, Italy, in collaboration with the Italian Space Agency (ASI), Rome, Italy. He was the winner of the IEEE GRS Italy Chapter Frank Marzano Award for one of the top three Master's theses in Geosciences and Remote Sensing. He has coauthored and presented papers at several prestigious conferences and received the Best Poster Award in Urban and Data Analysis as a young scientist at the 2024 European Space Agency (ESA)-Dragon Symposium. His research interests include the application of artificial intelligence and remote sensing to broader urban analysis tasks, with particular emphasis on damage mapping and exposure assessment following natural disasters and extreme events.
\end{IEEEbiography}
\end{minipage}

\begin{minipage}[t]{0.47\textwidth}
\begin{IEEEbiography}[{\includegraphics[width=1.05in,height=1.20in,clip,keepaspectratio]{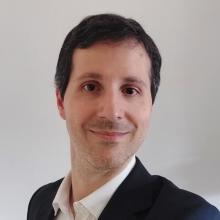}}]{Deodato Tapete} Researcher in Earth Observation and data analytics at the Italian Space Agency (ASI) since 2017, PhD in Earth Sciences, specialised in Synthetic Aperture Radar (SAR) and optical satellite remote sensing for monitoring of cultural heritage, archaeological remote sensing, assessment of natural and anthropogenic hazards, urban applications, agriculture and water resources management. He has developed several methods based on SAR data processing to address issues of heritage conservation, including but not limited to structural stability, impacts of infrastructure construction, weathering and deterioration, looting, intentional destruction. He is ASI program scientist for the Committee on Earth Observation Satellites (CEOS) Working Group on Disaster (WGD). He also leads the ASI programme "Innovation for Downstream Preparation for Science".
\end{IEEEbiography} 
\end{minipage}


\begin{minipage}[t]{0.47\textwidth}
\begin{IEEEbiography}[{\includegraphics[width=1.05in,height=1.20in,clip,keepaspectratio]{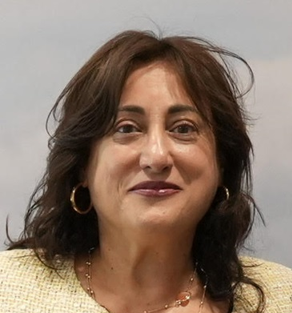}}]{Silvia Liberata Ullo}
IEEE Senior Member, President of IEEE AESS Italy Chapter, Industry Liaison for IEEE Joint ComSoc/VTS Italy Chapter since 2018, National Referent for FIDAPA BPW Italy Science and Technology Task Force (2019-2021). Member of the Image Analysis and Data Fusion Technical Committee (IADF TC) of the IEEE Geoscience and Remote Sensing Society (GRSS) since 2020. GRSS Europe Liaison since January 2024. Editor in Chief of IET IMage Processing. Graduated with laude in 1989 in Electronic Engineering at the University of Naples (Italy), pursued the M.Sc. in Management at MIT (Massachusetts Institute of Technology, USA) in 1992. Researcher and teacher since 2004 at University of Sannio, Benevento (Italy). Member of Academic Senate and PhD Professors’ Board. Courses: Signal theory and elaboration, Telecommunication networks (Bachelor program); Earth monitoring and mission analysis Lab (Master program), Optical and radar remote sensing (Ph.D. program). Authored 90+ research papers, co-authored many book chapters and served as editor of two books. Associate Editor of relevant journals (IEEE JSTARS, IEEE GRSL, MDPI Remote Sensing, IET Image Processing, Springer Arabian Journal of Geosciences and Recent Advances in Computer Science and Communications). Guest Editor of many special issues. Research interests: signal processing, radar systems, sensor networks, smart grids, remote sensing, satellite data analysis, machine learning and quantum ML applied to remote sensing.
\end{IEEEbiography} 
\end{minipage}


\begin{minipage}[t]{0.47\textwidth}
\begin{IEEEbiography}[{\includegraphics[width=1in,height=1.15in,clip,keepaspectratio]{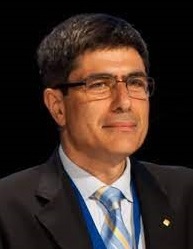}}]{Paolo Gamba}
IEEE Fellow, received the Laurea (cum laude) and Ph.D. degrees in electronic engineering from the University of Pavia, Pavia, Italy, in 1989 and 1993, respectively. He is a Professor of telecommunications with the University of Pavia, where he leads the Telecommunications and Remote Sensing Laboratory and serves as a Deputy Coordinator of the Ph.D. School in Electronics and Computer Science. He has been invited to give keynote lectures and tutorials in several occasions about urban remote sensing, data fusion, EO data, and risk management. Dr. Gamba has served as the Chair for the Data Fusion Committee of the IEEE Geoscience and Remote Sensing Society from 2005 to 2009. He has been elected in the GRSS AdCom since 2014. He was also the GRSS President. He had been the Organizer and Technical Chair of the biennial GRSS/ISPRS Joint Workshops on Remote Sensing and Data Fusion over Urban Areas from 2001 to 2015. He has also served as the Technical Co-Chair of the 2010, 2015, and 2020 IGARSS Conferences, Honolulu, HI, USA, and Milan, Italy, respectively. He was the Editor-in-Chief of the IEEE Geoscience and Remote Sensing Letters from 2009 to 2013.
\end{IEEEbiography}
\end{minipage}

\end{document}